\documentclass[10pt,twocolumn,letterpaper]{article}

\usepackage{cvpr}              

\usepackage{graphicx}
\usepackage{amsmath}
\usepackage{amssymb}
\usepackage{booktabs}

\usepackage{fixmath}
\usepackage{color}
\usepackage{multirow}
\usepackage{bm}
\usepackage{bbm}
\usepackage{caption}
\usepackage{subcaption}
\usepackage{epsfig}

\usepackage[pagebackref,breaklinks,colorlinks]{hyperref}

\usepackage[capitalize]{cleveref}
\crefname{section}{Sec.}{Secs.}
\Crefname{section}{Section}{Sections}
\Crefname{table}{Table}{Tables}
\crefname{table}{Tab.}{Tabs.}

\def\cvprPaperID{5616} 
\def\confName{CVPR}
\def\confYear{2023}

\begin{document}

\title{Multi-Modal Representation Learning with Text-Driven Soft Masks}

\author{Jaeyoo Park$^1$ \qquad\qquad\qquad Bohyung Han$^{1, 2}$ \\
Computer Vision Laboratory, ECE$^1$ \& IPAI$^{2}$, Seoul National University\\
{\tt\small \{bellos1203, bhhan\}@snu.ac.kr}
}

\maketitle


\begin{abstract}
We propose a visual-linguistic representation learning approach within a self-supervised learning framework by introducing a new operation, loss, and data augmentation strategy.
First, we generate diverse features for the image-text matching (ITM) task via soft-masking the regions in an image, which are most relevant to a certain word in the corresponding caption, instead of completely removing them.
Since our framework relies only on image-caption pairs with no fine-grained annotations, we identify the relevant regions to each word by computing the word-conditional visual attention using multi-modal encoder.
Second, we encourage the model to focus more on hard but diverse examples by proposing a focal loss for the image-text contrastive learning (ITC) objective, which alleviates the inherent limitations of overfitting and bias issues.
Last, we perform multi-modal data augmentations for self-supervised learning via mining various examples by masking texts and rendering distortions on images.
We show that the combination of these three innovations is effective for learning a pretrained model, leading to outstanding performance on multiple vision-language downstream tasks.

\end{abstract}
\vspace{-5mm}

\section{Introduction}
\label{sec:intro}

Vision-language representation learning aims at optimizing a joint embedding model for the data in both the image and text domains, where the learned representations are transferred to solve various vision-language downstream tasks including image-text retrieval~\cite{plummer2015flickr30k,lin2014microsoft,lee2021cosmo}, visual question answering~\cite{cadene2019murel,noh2016image,yu2019deep,anderson2018bottom}, visual reasoning~\cite{suhr2019corpus,xie2019visual}, etc.
With the introduction of large-scale image-text datasets~\cite{lin2014microsoft,krishna2017visual,ordonez2011im2text,sharma2018conceptual} and the recent advances in language models~\cite{devlin2019bert,brown2020language,vaswani2017attention}, research for vision-language representation learning~\cite{lu2019vilbert,chen2020uniter,li2020oscar,kim2021vilt,li2021align} has been actively exploring self-supervision tasks such as image-text matching (ITM), masked language modeling (MLM), and image-text contrastive learning (ITC).

\begin{figure}[t]
	\centering
	\includegraphics[width=\linewidth]{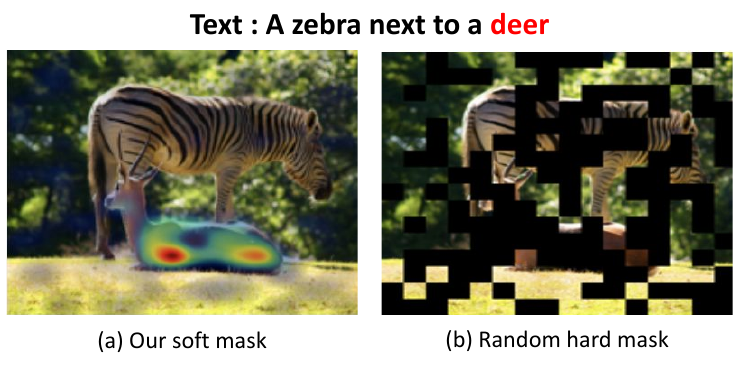}
	\vspace{-6mm}
	\caption{Illustration of the difference between the proposed soft feature masking and the random hard feature masking.
	The random hard feature masking has a risk of completely removing the important regions while our soft feature masking strategy still preserves the regions.}
	\vspace{-2mm}
\label{fig:softmask}
\end{figure}

During the pretraining stage, the model solves the tasks based only on image-caption pairs with no fine-grained annotations, \textit{i.e.,} annotations of region-word relations.
Thus, the model focuses on the most discriminative regions in images, which typically correspond to the primary objects in the scenes.
Hence, the model lacks a comprehensive understanding of various attributes observed in images despite detailed descriptions in captions.
To alleviate this limitation, we propose the following three techniques for diversifying observations, eventually resulting in performance improvement of learned representations.

First, we devise a soft masking technique on visual features for training, which suppresses activations at the discriminative parts in an input image with respect to a certain word in its caption.
The proposed soft feature masking makes a model focus on the regions that are initially paid less attention to and synthesizes diverse samples during the training procedure.
To be specific, we first utilize the cross-attention map of image and text features derived by the multi-modal fusion encoder and identify salient regions in the image by computing the word-specific Grad-CAM~\cite{selvaraju2017grad} of the image-text matching score.
Then we feed a softly-masked image feature to the fusion encoder for learning to match the features of both modalities.
While the most straightforward way to augment diverse visual features is to employ discrete hard masks as recent trends~\cite{he2022masked,xie2022simmim}, which aim to reconstruct images, our approach suppresses information in images using the soft masks with real-valued weights instead of completely removing the content relevant to input text.
Figure~\ref{fig:softmask} illustrates the difference between the proposed soft mask and the random hard (discrete) mask.

Second, we also encourage the model to focus more on hard examples by adopting a focal version of the ITC loss.
The focal loss~\cite{lin2017focal} is originally designed for object detection to prevent overfitting and class imbalance problems, and a similar idea turns out to be effective for our contrastive learning task.
This is partly because our task still has overfitting issues for easy examples and is also biased toward negative examples.
Since large-scale image-caption corpora are composed of multiple datasets with large domain gaps, a model distinguishes a lot of samples from different datasets very easily~\cite{cui2022contrastive}.
Hence, we propose the focal ITC loss for our model to achieve more balanced training.

In addition to the two components discussed above, we perform multi-modal data augmentations, which include diverse strong augmentations proposed in~\cite{yang2022vision} and binary masks to captions for the ITM task. 
Strong data augmentations have been believed to harm vision-language models~\cite{tang2020semantic,kant2021contrast} by breaking the semantic consistency of the image-text pair, but we argue that this does not hold for the large-scale vision-language representation pretraining, as also shown in recent studies~\cite{liang2022mind,jiang2023understanding}.
Note that the use of binary masks in ITM also homogenizes text inputs with the masked language model (MLM) task.

The organization of this paper is as follows.
We first discuss related works about vision-language representation learning in Section~\ref{sec:related} and provide the preliminaries about vision-language representation learning in Section~\ref{sec:preliminaries}. 
Section~\ref{sec:method} discusses the vision-language representation learning approach equipped with our ideas.
We demonstrate the performance of the proposed approach on multiple downstream tasks in Section~\ref{sec:experiments} and conclude this paper in Section~\ref{sec:conclusion}.


\section{Related Works}
\label{sec:related}

Vision-language representation learning approaches are mainly split into the methods equipped with the pretrained object detector~\cite{lu2019vilbert,tan2019lxmert,chen2020uniter,li2020oscar,gan2020large,zhang2021vinvl} and the ones with a detector-free visual encoder~\cite{kim2021vilt,huang2021seeing,li2021align,yang2022vision,duan2022multi}.
Among lots of literature about the topic, below we provide the ones that are most related to ours. 

Following the studies for various vision-language tasks such as visual question answering~\cite{cadene2019murel,yu2019deep,anderson2018bottom} and image-text matching~\cite{lee2018stacked,li2019visual}, research in the early day adopted a bottom-up strategy with a pretrained object detector, \eg, Faster R-CNN~\cite{ren2015faster}, as a visual encoder.
ViLBERT~\cite{lu2019vilbert} and LXMERT~\cite{tan2019lxmert} directly extend BERT~\cite{devlin2019bert} by constructing co-attention-based frameworks with two separate modality-specifc streams.
There exist some approaches deploying single-stream transformers as fusion encoders. 
UNITER~\cite{chen2020uniter} proposes word-region alignment loss and investigates various mask modeling objectives.
Meanwhile, OSCAR~\cite{li2020oscar} incorporates object tags from detection results as anchors to reduce the gap between image and text embedding spaces.
Based on~\cite{chen2020uniter}, \cite{gan2020large}~adopts an adversarial training strategy to improve generalization capability, although being yet time-consuming.
Recently, \cite{zhang2021vinvl}~studies the effect of an improved object detector and provides rich object embeddings.

However, approaches incorporating fixed pretrained object detectors as their visual encoders have several drawbacks. 
First, the visual encoders cannot be trained during training, due to the high computation cost.
Although detector-based visual encoders provide rich embeddings for the pre-defined object categories, they suffer from the lack of information about the objects that are not seen during the training of the detection model. 
Second, the visual embeddings are learned from bounding box regions focusing on objects.
Hence, the reasoning process fails to learn the global context from the outside of the bounding boxes, such as the background. 

\begin{figure*}[t!]
	\centering
	\includegraphics[width=0.9\linewidth]{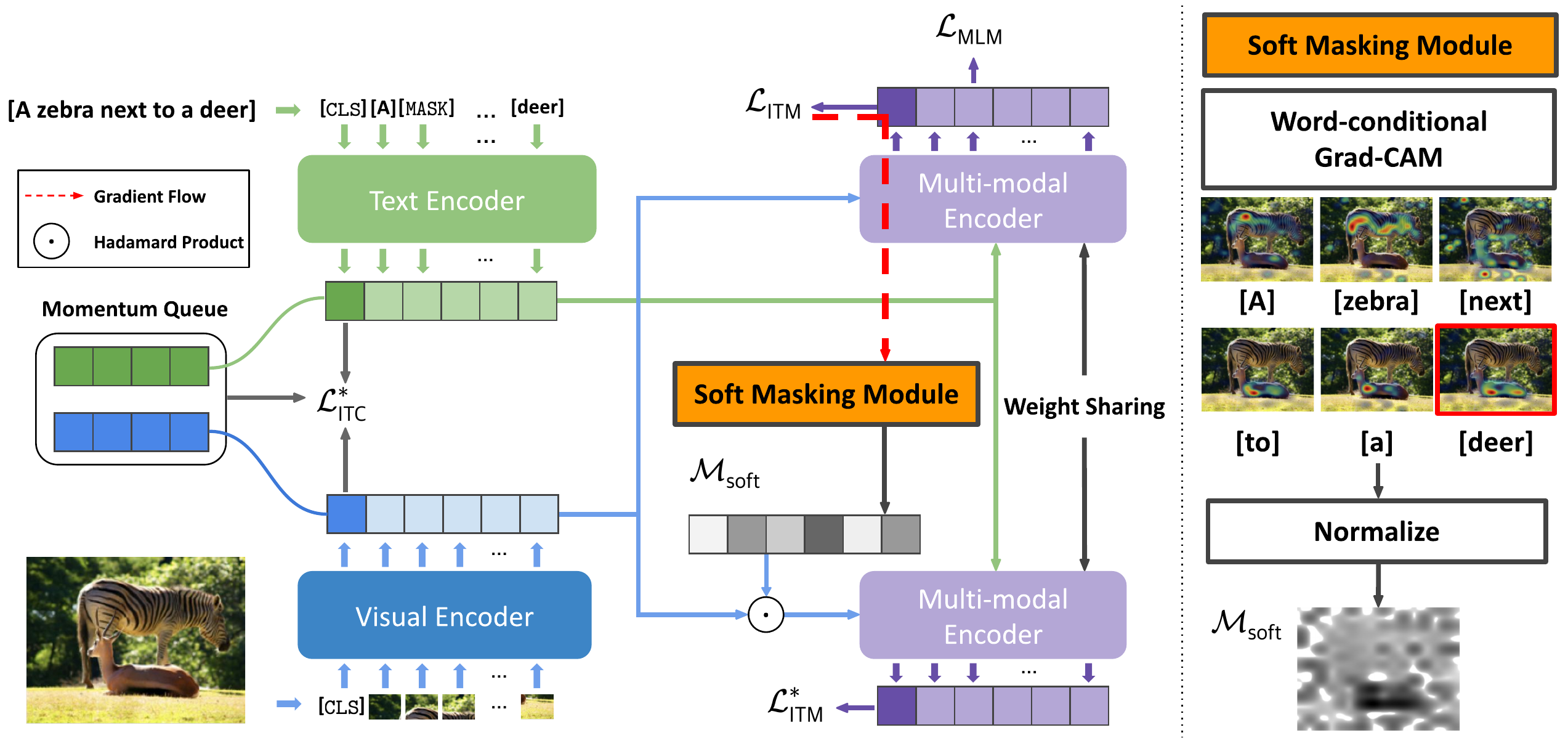}
	\caption{Illustration of the proposed framework.
	Given the image and the masked text data, we encode them to obtain the sequences of unimodal embedding tokens from both modalities.
	Each encoder has a paired momentum model updated by the moving-average, which we omit in the figure for simplicity.
	For the unimodal level, the model optimizes ITC objective to align the embedding space of each modality before fusing them (Section~\ref{sec:preliminaries}).
	We feed the sequences of embedding tokens to the multi-modal encoder to aggregate the information of both modalities and learn to solve ITM and MLM tasks (Section~\ref{sec:preliminaries}).
	We propose to additionally employ soft masked visual features for learning complement attributes from the image (Section~\ref{sec:soft_masking}).
	We generate the soft masks from the Grad-CAM of the positive matching score with respect to the cross-attention map of the multi-modal encoder. 
	We randomly select one of the word-wise Grad-CAM and use the normalized Grad-CAM as a soft mask for the visual feature.
	Our generated soft-masked visual feature works as a hard but diversified sample by softly-masking the important regions while not perfectly removing them.
	Moreover, we introduce the focal version of ITC, where the model focuses more on the hard examples (Section~\ref{sec:focal_itc}). 	 
	} 
	\vspace{-2mm}
\label{fig:framework}
\end{figure*}

To address such drawbacks, recent works attempt to build models with detector-free visual encoders.
For example, VilT~\cite{kim2021vilt}~feeds the linear projections of grid-patch embeddings within an image, but shows lower performance due to the lack of the deep visual encoder.
SOHO~\cite{huang2021seeing} discretizes the embedding of an input image using an end-to-end trainable CNN-based visual encoder with a visual dictionary.
ALBEF~\cite{li2021align}~adopts a Vision Transformer~\cite{dosovitskiy2021image} and proposes to align the intermediate features of two modalities before being fused, using contrastive objective.
TCL~\cite{yang2022vision}~further learns to maximize the mutual information between local regions and global embeddings in each modality, based on the features from~\cite{li2021align}.
CODIS~\cite{duan2022multi}~proposes a multimodal codebook learning, and bridges the gap between two modalities. 
While our work shares the basic architecture with \cite{li2021align}~and~\cite{yang2022vision}, its main goal is to take advantage of softly masked visual features and incorporate various regularization techniques through the construction of diversified examples.


\section{Preliminaries}
\label{sec:preliminaries}
This section discusses the details of baseline architectures and self-supervised learning tasks, which are widely used for vision-language representation learning.

\subsection{Multi-Modal Representation Learning}
\label{sec:multi}
The model consists of a visual encoder $f_v$, a text encoder $f_t$, and a multi-modal encoder $f_{mm}$.
Each encoder has a paired momentum encoder and the model stores historical \texttt{[CLS]} tokens for training.
An input image $I$ is first patchified into a sequence of small grid patch tokens, $\Vec{I}$.
We encode a set of image patch tokens, $\Vec{I}$, into a sequence of embedding tokens using an image encoder, $f_v$, and obtain $\Vec{V}_\text{emb}=\{\Vec{v}_0, \Vec{v}_1,...,\Vec{v}_N\}$, where $\Vec{v}_0$ denotes the \texttt{[CLS]} token feature and $N$ is number of the patches in the input image.
Likewise, we tokenize an input text $T$ into a sequence of word tokens $\Vec{T}$, and the text encoder embeds $\Vec{T}$ into a sequence of text features $\Vec{T}_\text{emb}=\{\Vec{t}_0, \Vec{t}_1,...,\Vec{t}_L\}$, where $\Vec{t}_0$ denotes the \texttt{[CLS]} token feature and $L$ is the length of the input text.
The image features $\Vec{V}_\text{emb}$ are fused with the text features $\Vec{T}_\text{emb}$ through a cross-attention at each layer of the multi-modal encoder.
The multi-modal encoder, $f_{mm}$, outputs the joint embedding tokens of two modalities, $\Vec{M}_\text{emb}=\{\Vec{m}_0, \Vec{m}_1,...,\Vec{m}_L\}$, where $\Vec{m}_0$ denotes the \texttt{[CLS]} token feature. 
Note that $|\Vec{T}_\text{emb}| = |\Vec{M}_\text{emb}| = L + 1$, since the multi-modal encoder takes $\Vec{T}_\text{emb}$ as its query and fuses $\Vec{V}_\text{emb}$ and $\Vec{T}_\text{emb}$ via a cross-attention mechanism.
Based on this configuration, we perform multi-modal self-supervised representation learning by using the following objectives.

\subsection{Self-Supervised Learning Objectives}
We discuss three tasks based on self-supervision, which are widely used for multi-modal representation learning.

\vspace{-3mm}
\paragraph{Image-Text Contrastive Learning (ITC)}
ITC aligns two modality-specific representations, \ie, \texttt{[CLS]} tokens denoted by $\Vec{v}_{0}$ and $\Vec{t}_{0}$ for an image and a text, before their aggregations using a multi-modal fusion encoder. 
In other words, ITC enforces an image-text pair, $(\Vec{v}_{0}, \Vec{t}_{0})$, to be close if they are matched, and to be far away otherwise.
Following ALBEF~\cite{li2021align}, we adopt two momentum queues for storing recent \texttt{[CLS]} token features from two unimodal encoders, \textit{i.e.,} $\Vec{v}_{0}$ and $\Vec{t}_{0}$, and use them as additional examples for contrastive learning. 
Formally, the similarity between $I$ and $T$, $s(I,T)$, is defined as
\begin{align}
s(I,T) = h_v(\Vec{v}_{0})^{T}h_t(\Vec{t}_{0}),
\end{align}
where $h_v(\cdot)$ and $h_t(\cdot)$ are linear projection layers.
Then, for each $I$ and $T$, the image-to-text similarity score for the $i^\text{th}$ example is defined as 
\begin{align}
    p_{v2t}^{(i)} = \frac{\exp(s(I_i,T_i)/\tau)}{\sum_{j=1}^{B+Q}\exp(s(I_i,T_j)/\tau)},
    \label{eq:itc_similarity}
\end{align}
where $\tau$ is a learnable temperature, $B$ denotes the size of mini-batch, and $Q$ is the size of the momentum queue.
Note that $\{T_j | j = B+1, ..., B+Q\}$ is a set of the negative text samples stored in the momentum queue.  
We also define $p_{t2v}^{(i)}$, text-to-image similarity score for the $i^\text{th}$ data, in the same way. 
The ITC loss, which is the sum of InfoNCE losses~\cite{oord2018representation} in two directions, is then given by
\begin{align}
\label{eq:itc_loss}
   &\mathcal{L}_\text{ITC} =  -\frac{1}{2B} \sum_{i=1}^{B} \left[\log p_{v2t}^{(i)} + \log p_{t2v}^{(i)} \right].
\end{align}

\vspace{-3mm}
\paragraph{Image-Text Matching (ITM)}
The ITM task aims to classify whether a given image and text pair is matched or not.
The classification score of the $i^\text{th}$ image-text pair in a mini-batch, $\bm{q}_\text{ITM}^{(i)} \in \mathbb{R}^2$, is obtained by feeding the final joint embedding feature, $\Vec{m}_0$, to a binary linear classifier $h_{itm}(\cdot)$, which is given by
\begin{equation}
  \bm{q}_\text{ITM}^{(i)} = h_{itm}(\Vec{m}_0^{(i)})=h_{itm}(f_{mm}(\Vec{V}_\text{emb}^{(i)},\Vec{T}_{emb}^{(i)})[0]).
   \label{eq:itm_pred}
\end{equation}
The ITM loss is defined as 
\begin{align}
  & \mathcal{L}_\text{ITM} = \frac{1}{S} \sum_{i=1}^{S} H \left( \bm{y}_{\text{ITM}}^{(i)},\bm{q}_\text{ITM}^{(i)} \right),
   \label{eq:itm_loss}
\end{align}
where $\bm{y}_\text{ITM}^{(i)}$ is a one-hot vector representing the ground-truth label, $S$ denotes the number of image-text pairs, and $H(\cdot, \cdot)$ stands for the cross-entropy function.

\vspace{-3mm}
\paragraph{Masked Language Modeling (MLM)}
MLM aims to predict the masked text by employing both the image and the contextualized text.
Following BERT \cite{devlin2019bert}, the text tokens of the $i^\text{th}$ text from the image-text pair, $\Vec{T}^\text{(i)}$, are randomly masked out with a probability of 15\%, and then the special token \texttt{[MASK]} replaces the selected tokens, which results in $\hat{\Vec{T}}^{(i)}$. 
Given the output of multi-modal encoder, the probability of a masked token over the predefined vocabulary, $\Vec{q}_{\text{MLM}}^{(i)} \in \mathbb{R}^{30522}$ is obtained by feeding the joint embedding feature of the masked text token, $\Vec{m}_\text{mask}^{(i)} \in \Vec{M}_\text{emb}^{(i)}$, to a linear classifier $h_{mlm}(\cdot)$ as
\begin{equation}
  \bm{q}_\text{MLM}^{(i)} = h_{mlm}(\Vec{m}_\text{mask}^{(i)}). 
   \label{eq:mlm_pred}
\end{equation}
Then, the MLM objective is defined as follows: 
\begin{equation}
    \mathcal{L}_{\text{MLM}} = \frac{1}{S} \sum_{i=1}^{S} \mathbb{E}_{(\Vec{I}, \hat{\Vec{T}}) \sim \mathcal{D}}\left[ H \left( \bm{y}_{\text{MLM}}^{(i)},\bm{q}_{\text{MLM}}^{(i)} \right) \right].
\end{equation}


\section{Method}
\label{sec:method}
This section describes the proposed approach, including text-driven soft feature masking, focal image-text contrastive learning, and multi-modal data augmentation. 
Figure~\ref{fig:framework} illustrates the overall procedure and network architecture of our approach.

\subsection{Text-Driven Soft Feature Masking}
\label{sec:soft_masking}
Our model is pretrained based only on image-caption pairs and is prone to overfit discriminative regions in images as often observed in image recognition tasks.
Hence, the model may not be able to understand the details of input images despite the availability of matched captions.

To tackle this issue, we propose to learn multi-modal representations by matching the masked image embedding and the corresponding text embeddings. 
Contrary to a common hard masking strategy, we generate soft masks by using the concept of Grad-CAM~\cite{selvaraju2017grad}, which highlights the relevant parts in an image on the output of the learned model.
Because it is critical to maintain crucial information in an input image for accurate matching between images and texts while augmenting hard examples by partly masking informative regions, we believe that soft masking based on a word-conditional Grad-CAM is a reasonable idea and adopt it as a new operation in our algorithm.
As shown in the right-hand side of Figure~\ref{fig:framework}, the word-conditional Grad-CAM effectively captures the discriminative regions, which correspond to the objects in general, even for stop-words.

Specifically, for the $i^\text{th}$ image-text pair in a mini-batch, we first identify salient regions in the image by computing the Grad-CAM of the initial image-text matching score, $\Vec{q}_\text{ITM}^{(i)}$, using the cross-attention map of the image and text embeddings given by the multi-modal fusion encoder.
Let $A_k^{(i)}\in\mathbb{R}^{(L+1)\times(N+1)}$ denote the cross-attention map in the $k^\text{th}$ transformer block of the multi-modal encoder. 
We obtain the Grad-CAM denoted by $A_\text{GCAM}^{(i)}\in\mathbb{R}^{(L+1)\times(N+1)}$ of the positive matching score with respect to $A_k^{(i)}$ as
\begin{equation}
\label{eqn:grad_cam}
A_\text{GCAM}^{(i)} =  \frac{1}{K}\sum_{k=1}^{K}\text{ReLU} \left(\frac{\partial q_\textrm{ITM}^{+(i)}}{\partial A_k^{(i)}} \odot A_k^{(i)} \right),
\end{equation}
where $q_\textrm{ITM}^{+(i)}$ denotes the positive matching score of the ITM task, which is equal to $\bm{q}_\textrm{ITM}^{(i)}[1]$.
Note that we aggregate the Grad-CAM maps from all cross-attention layers to obtain a more accurate Grad-CAM.

Among $L+1$ word tokens, including the \texttt{[CLS]} token, we randomly sample a single word index from a uniform distribution and obtain the corresponding word-conditional Grad-CAM.
The rationale behind the random selection is to boost stochasticity and obtain attention for diverse regions during training.
Let $i_w$ be the index of the sampled word. 
Then, the text-driven soft mask for an image embedding, $\mathcal{M}_\text{soft}^{(i)} \in\mathbb{R}^{N+1}$, is given by
\begin{align}
\label{eqn:mask}
\mathcal{M}_\text{soft}^{(i)} = \mathbbm{1} - \hat{A}_\text{GCAM}^{(i)}[i_w], 
\end{align}
where $\hat{A}_\text{GCAM}^{(i)}[i_w]$ is the normalized value of $A_\text{GCAM}^{(i)}[i_w]$ with min-max clipping.
We compute the masked embedding $\hat{\Vec{V}}_\text{emb}^{(i)}$, which is given by
\begin{equation}
\quad \hat{\Vec{V}}_\text{emb}^{(i)} = \mathcal{M}_\text{soft}^{(i)}  \odot \Vec{V}_\text{emb}^{(i)},
\label{eq:masked_embedding}
\end{equation}
and feedforward it to the multi-modal encoder to obtain new multi-modal joint embedding, $\hat{\Vec{M}}_\text{emb}^{(i)}$.
By classifying $\hat{\Vec{m}}_0^{(i)}$ with $h_{itm}$, we obtain the prediction scores for the masked soft feature as
\begin{align}
\hat{\bm{q}}_\text{ITM}^{(i)} = h_{itm} (\hat{\Vec{m}}^{(i)}_0)=h_{itm} \left( f_{mm} ( \hat{\Vec{V}}_\text{emb}^{(i)},\Vec{T}_\text{emb}^{(i)})[0] \right).
\end{align}
Then, the ITM loss of the soft masked feature is defined as 
\begin{equation}
\mathcal{L}^*_\text{ITM} = \frac{1}{S'}\sum_{i=1}^{S'}H(\Vec{y}_\text{ITM}^{(i)}, \hat{\Vec{q}}_\text{ITM}^{(i)}), 
\label{eq:itm_soft}
\end{equation}
where $\Vec{y}_\text{ITM}^{(i)}$ is the same ground-truth vector used in~\eqref{eq:itm_loss} and $S'$ is the number of positive pairs in a mini-batch.

\subsection{Focal Image-Text Contrastive Learning}
\label{sec:focal_itc}
To boost the regularization effect from hard examples, we design a variant of the ITC loss, which is similar to the focal loss~\cite{lin2017focal}. 
We define the focal ITC loss as
\begin{align}
    \label{eq:focal_ITC_loss}
   & \mathcal{L}^*_\text{ITC} =  \\
    & -\frac{1}{2B} \sum_{i=1}^{B} \left[ (1-p_{v2t}^{ (i) } )^{\gamma} \log p_{v2t}^{(i)} + (1-p_{t2v}^{(i)})^{\gamma}  \log p_{t2v}^{(i)} \right], \nonumber
\end{align}
where $\gamma$ is a modulating factor to reduce weights for easy samples and is set to $2$ following the original version of the focal loss.  
The proposed focal ITC loss alleviates the overfitting to easy examples while handling the class imbalance issue effectively.
Since large-scale image-caption corpora are composed of multiple datasets with large domain gaps, a model distinguishes a lot of samples from different datasets easily, as also revealed in~\cite{cui2022contrastive}.
Hence, in this sense, the focal ITC loss is particularly useful for our training scheme.

\subsection{Multi-Modal Data Augmentation}
We perform multi-modal data augmentations (MMDAs), which include diverse strong augmentations proposed in~\cite{yang2022vision} and binary masking to captions for the ITM task. 
We propose to generate hard positive samples by applying strong augmentations on both modalities.
We crop images randomly to resolution $256 \times 256$ and apply RandAugment~\cite{cubuk2020randaugment}.
We also perform color distortions such as random color jittering, random grayscale conversion, and random Gaussian blur.
On the text side, we replace $\Vec{T}$ with $\hat{\Vec{T}}$. 
As illustrated in Section~\ref{sub:ablation}, such a simple augmentation strategy turns out to improve performance despite the preconception that it deteriorates semantic relations on vision-language tasks.
Note that the data augmentation also affects the ITC.
We reduce the computational cost of the whole framework because we now perform forward the text encoder just once unlike the previous works~\cite{li2021align,yang2022vision}\footnote{While the papers described their models as if they learn ITM with augmented images and masked texts, we found that their ITM takes text without masks and they have to feedforward text features twice with and without masks because MLM adopts masked text as its input.}.
Such the saved cost allows us to compute another forward pass for soft masking and optimizing $\mathcal{L}_\text{ITM}^*$ in \eqref{eq:itm_soft}. 
The analysis on computation cost is presented in Section~\ref{sub:cost}.

\subsection{Training Objective}
\label{sec:method_objective}
The final objective function of our algorithm, denoted by $\mathcal{L}_{\text{Final}}$, is given by
\begin{equation}
\mathcal{L}_{\text{Final}} = \mathcal{L}_\text{ITM}+ \mathcal{L}_\text{ITC}^* + \mathcal{L}_\text{MLM} + \mathcal{L}_\text{ITM}^*. 
\label{eq:final_loss_definition}
\end{equation}

\subsection{Discussion}
Our work introduces the aforementioned three components to utilize diversely augmented image-text pairs and facilitate large-scale multi-modal pretraining. 
We employ a novel text-driven soft feature masking strategy to populate examples with diverse aspects, where the discriminative parts of an image are softly masked, at the intermediate feature level.
By feeding the features covered by soft masks, the model learns to see complementary regions without completely sacrificing major information whereas conventional hard masking often loses critical contents.
Next, we encourage the model to pay more attention to the hard examples, by adopting a focal version of the image-text contrastive learning objective.
Last, we augment data in the input level and make the model learn with hard image-text pairs.
Although employing imperfectly aligned image-text pairs is counter-intuitive and not explored much, our ideas align with recent studies~\cite{liang2022mind,jiang2023understanding} that show perfect modality alignment does not guarantee performance improvement in downstream task.

\vspace{-3mm}


\begin{table*}[t!]
	\caption{Performance comparison of fine-tuned image-text retrieval on the Flickr30K and MS-COCO datasets. 
	Our approach is named as SoftMask++: SoftMask together with Focal ITC and MMDA.
	For completeness, we also provide the results of ALIGN \cite{jia2021scaling}, which uses 1.8B image-text pairs (1.2B unique images) for pretraining.
	The bold-faced numbers indicate the best performance.}
	\vspace{-0.5cm}
	\setlength\tabcolsep{7.5pt}
	\begin{center}
	\scalebox{0.85}{
		\begin{tabular}{l|c|cccccc|cccccc}
            \toprule 
                \multirow{3}{*}{Method} &
                \multirow{3}{*}{\#Img} & \multicolumn{6}{c}{Flickr30K (1K)} & \multicolumn{6}{c}{MS-COCO (5K)}\\
                
                & & \multicolumn{3}{c}{Text Retrieval} & \multicolumn{3}{c}{Image Retrieval} &
                \multicolumn{3}{c}{Text Retrieval} & \multicolumn{3}{c}{Image Retrieval} \\
            
                & & R@1 & R@5 & R@10 & R@1 & R@5 & R@10
                & R@1 & R@5 & R@10 & R@1 & R@5 & R@10 \\
                \midrule
                UNITER \cite{chen2020uniter} & 4M & 87.3 & 98.0 & 99.2 & 75.6 & 94.1 & 96.8 & 65.7 & 88.6 & 93.8 & 52.9 & 79.9 & 88.0  \\
                VILLA \cite{gan2020large} & 4M & 87.9 & 97.5 & 98.8 & 76.3 & 94.2 & 96.8 & --- & --- & --- & --- & --- & ---  \\
                OSCAR \cite{li2020oscar} & 4M & --- & --- & --- & --- & --- & --- & 70.0 & 91.1 & 95.5 &  54.0 & 80.8 & 88.5  \\
                ViLT \cite{kim2021vilt} & 4M & 83.5 & 96.7 & 98.6 &  64.4 & 88.7 & 93.8 & 61.5 & 86.3 & 92.7  & 42.7 & 72.9 & 83.1  \\
                UNIMO \cite{li2020unimo} & 4M &  89.7 & 98.4 & 99.1 & 74.7 & 93.5 & 96.1 & --- & --- & --- & --- & --- & ---  \\
                SOHO \cite{huang2021seeing} & 200K & 86.5 & 98.1 & 99.3 &  72.5 & 92.7 & 96.1 & 66.4 & 88.2 & 93.8 & 50.6 & 78.0 & 86.7 \\
                ALBEF \cite{li2021align} & 4M & 94.3 & 99.4 & 99.8 & 82.8 & 96.7 & 98.4 & 73.1 & 91.4 & 96.0 & 56.8 & 81.5 & 89.2 \\                
                TCL \cite{yang2022vision} & 4M & 94.9 & 99.5 & 99.8 & 84.0 & 96.7 & \textbf{98.5} & 75.6 & 92.8 & \textbf{96.7} & 59.0 & 83.2 & 89.9 \\
                CODIS \cite{duan2022multi} & 4M & 95.1 & 99.4 & \textbf{99.9} & 83.3 & 96.1 & 97.8 & 75.3 & 92.6 & 96.6 & 58.7 & 82.8 & 89.7 \\
                VinVL \cite{zhang2021vinvl} & 6M & --- & --- & --- & --- & --- & --- & 75.4 & 92.9 & 96.2 & 58.8 & 83.5 & 90.3  \\
                \hline
	        SoftMask++ (ours) & 4M & \textbf{95.4} & \textbf{99.7} & \textbf{99.9} & \textbf{84.6} & \textbf{96.8} & \textbf{98.5} & \textbf{76.6} & \textbf{93.5}  & 96.6 & \textbf{60.2}  & \textbf{83.7} & \textbf{90.5} \\   
	        \hline           
                ALIGN \cite{jia2021scaling} & 1.2B & 95.3 & 99.8 & 100.0 & 84.9 & 97.4 & 98.6 & 77.0 & 93.5 & 96.9 & 59.9 & 83.3 & 89.8 \\

                \bottomrule
            \end{tabular}
            }
	\end{center}
	\vspace{-2mm}
	\label{table:fine_tune_itr}
\end{table*}

\begin{table*}
	\caption{Performance comparison of zero-shot image-text retrieval on the Flickr30K and MS-COCO datasets. 
	Our approach is named as SoftMask++: SoftMask together with Focal ITC and MMDA.
	For completeness, we also provide the results of ALIGN \cite{li2021align}, which uses 1.8B image-text pairs (1.2B unique images) for pretraining.
	The bold-faced numbers indicate the best performance.}
	\vspace{-0.5cm}
	\setlength\tabcolsep{7.5pt}
	\begin{center}
	\scalebox{0.85}{
		\begin{tabular}{l|c|cccccc|cccccc}
            \toprule 
                \multirow{3}{*}{Method} & \multirow{3}{*}{\#Img} & \multicolumn{6}{c}{Flickr30K (1K)} & \multicolumn{6}{c}{MS-COCO (5K)} \\
                
                & & \multicolumn{3}{c}{Text Retrieval} & \multicolumn{3}{c}{Image Retrieval} &
                \multicolumn{3}{c}{Text Retrieval} & \multicolumn{3}{c}{Image Retrieval} \\
            
                & & R@1 & R@5 & R@10 & R@1 & R@5 & R@10
                & R@1 & R@5 & R@10 & R@1 & R@5 & R@10 \\
                \midrule
                UNITER \cite{chen2020uniter} & 4M & 80.7 & 95.7 & 98.0 & 66.2 & 88.4 & 92.9  & 64.1 & 87.7 & 93.3 & 48.8 & 76.7 & 85.8 \\
                ViLT \cite{kim2021vilt} & 4M & 73.2 & 93.6 & 96.5 &  55.0 & 82.5 & 89.8 & 56.5 & 82.6 & 89.6 &  40.4 & 70.0 & 81.1 \\
                CLIP \cite{radford2021learning} & 400M & 88.0 & 98.7 & 99.4 & 68.7 & 90.6 & 95.2 & 58.4 & 81.5 & 88.1 & 37.8 & 62.4 & 72.2 \\
                ALIGN \cite{jia2021scaling}& 1.2B & 88.6 & 98.7 & 99.7 & 75.7 & 93.8 & 96.8 & 58.6 & 83.0 & 89.7 & 45.6 & 69.8 & 78.6 \\
                ALBEF \cite{li2021align} & 4M & 90.5 & 98.8 & 99.7 & 76.8 & 93.7 & 96.7 & 68.7 & 89.5 & 94.7 & 50.1 & 76.4 & 84.5 \\                
                TCL \cite{yang2022vision} & 4M & 93.0 & 99.1 & 99.6 & 79.6 & \textbf{95.1} & 97.4 & 71.4 & 90.8 & 95.4 & 53.5 & 79.0 & 87.1 \\
                CODIS \cite{duan2022multi} & 4M & 91.7 & \textbf{99.3} & \textbf{99.8} & 79.7 & 94.8 & 97.3 & 71.5 & 91.1 & 95.5 & 53.9 & 79.5 & 87.1 \\
                \hline
                SoftMask++ (ours) & 4M & \textbf{93.4}  & \textbf{99.3} & \textbf{99.8} & \textbf{80.1}  & 94.9 & \textbf{97.7}  & \textbf{72.3} & \textbf{91.5} & \textbf{95.7}  & \textbf{54.1}  & \textbf{79.8}  & \textbf{87.3}  \\
                \bottomrule
            \end{tabular}
            }
	\end{center}
	\vspace{-2mm}
	\label{table:zero_shot}
\end{table*}

\section{Experiments}
\label{sec:experiments}
This section presents the experimental results of our approach, referred to as SoftMask++, on various downstream tasks.
We also demonstrate the effectiveness of our ideas via several ablation studies.

\subsection{Pretraining Datasets} 
\label{sub:datasets}
We utilize large-scale image-text corpora for pretraining, which include MS-COCO~\cite{lin2014microsoft}, Visual Genome~\cite{krishna2017visual}, SBU Captions~\cite{ordonez2011im2text}, and Conceptual Captions~\cite{sharma2018conceptual}.
The combined dataset consists of 4M images and 5.1M image-text pairs as in the one used for previous works~\cite{li2021align,yang2022vision,duan2022multi,chen2020uniter}.

\subsection{Downstream Tasks}
\label{sub:downstream}
We briefly describe several vision-language downstream tasks used for the evaluation of our pretraining approach.
Refer to the supplementary document for more details.

\vspace{-3mm}
\paragraph{Image-Text Retrieval (ITR)}
We assess the effectiveness of our method on Image-Text Retrieval task, which consists of two subtasks, which are image-to-text retrieval (TR) and text-to-image retrieval (IR).
We evaluate the proposed approach on the Flickr30K~\cite{plummer2015flickr30k} and MS-COCO~\cite{lin2014microsoft} benchmarks using two different settings following the previous protocol~\cite{duan2022multi,li2021align,yang2022vision}.
In the first setting, we fine-tune the pretrained model using the training data in each dataset.
We also report the zero-shot performance, where the pre-trained model is directly evaluated on the test data without fine-tuning.
Note that, for zero-shot evaluation on Flickr30K, we exploit the model fine-tuned on MS-COCO, following the convention~\cite{duan2022multi,li2021align,yang2022vision}.

\vspace{-3mm}
\paragraph{Visual Entailment (VE)}
This is a fine-grained visual reasoning task that aims to predict whether an image semantically entails, contradicts, or is neutral to a text.
We use the SNLI-VE~\cite{xie2019visual} dataset, which consists of 30K, 1K, and 1K images as training, validation, and testing sets, respectively.
We predict the probability by feeding the joint embedding, $m_0$, to a linear classifier.

\vspace{-3mm}
\paragraph{Natural Language Visual Reasoning (NLVR)}
This task learns to reason about whether the text description is relevant to a pair of input images.
Since the task requires two images as inputs, we modify our model by duplicating transformer blocks following~\cite{li2021align,yang2022vision} and perform an additional pretraining step to accommodate the multi-modal encoder for the image pair.
We evaluate our model on the NLVR$^2$ dataset, which consists of 86K, 7K, and 7K examples for training, development, and test splits, respectively.

\vspace{-3mm}
\paragraph{Visual Question Answering (VQA)}
Visual question answering (VQA) is a task to find an answer for a question about an input image, which requires understanding local and global context of an image as well as a question.
We conduct experiments on VQA2.0 dataset~\cite{goyal2017making}, which is based on the images collected from the MS-COCO dataset.
We formulate the VQA task as an answer generation problem, following~\cite{li2021align,yang2022vision}.
The answer decoder is initialized with the pretrained weights from the multi-modal encoder and is fine-tuned on the training datasets to generate answers from the pre-defined candidates.

\begin{table}
	\caption{Performance comparison on various vision-language tasks, including VE, NLVR$^2$, and VQA.
	The bold-faced numbers indicate the best performance.
	Note that O.D. denotes object detector and the methods with the remark exploit the pretrained object detector as visual encoders, and D.F. represents the detector-free models.}
	\vspace{-0.5cm}
	\setlength\tabcolsep{2.5pt}
	\begin{center}
	\scalebox{0.85}{
		\begin{tabular}{clccccccc}
            \toprule 
             &
                \multicolumn{1}{l}{\multirow{2}{*}{Method}} &
                \multirow{2}{*}{\#Img} & \multicolumn{2}{c}{SNLI-VE} & \multicolumn{2}{c}{NLVR$^2$} &
                \multicolumn{2}{c}{VQA}\\
                
                & & & val & test & dev & test-P & dev& std \\
                \midrule
                \multirow{5}{*}{O.D.}&OSCAR \cite{li2020oscar} & 4M & ---  & --- & 78.1 & 78.4 & 73.2 & 73.4 \\

                &UNITER \cite{chen2020uniter} & 4M & 78.6 & 78.3 & 77.2 & 77.9 & 72.7 & 72.9  \\
                
                &UNIMO \cite{li2020unimo} & 4M & 80.0 & 79.1 & --- & --- & 73.3 & 74.0 \\
                
                &VILLA \cite{gan2020large} & 4M & 79.5 & 79.0 & 78.4 & 79.3 & 73.6 & 73.7 \\
                
                &VinVL \cite{zhang2021vinvl} & 6M & --- & --- & 82.1 & 83.1 & 75.9 & 76.1 \\ \midrule
                
               \multirow{5}{*}{D.F.} &ViLT \cite{kim2021vilt} & 4M & --- & --- & 75.7 & 76.1 & 71.3 & --- \\
                
                &ALBEF \cite{li2021align} & 4M & 80.1 & 80.3 & 80.2 & 80.5 & 74.5 & 74.7  \\
                
                &TCL \cite{yang2022vision}   & 4M & 80.5 & 80.3 & 80.5 & 81.3 & 74.9 & 74.9 \\
               
                &CODIS \cite{duan2022multi} & 4M & 80.5  & 80.4 & 80.5 & 80.8 &\textbf{75.0} &74.9 \\
                
                \cmidrule(lr){2-9}
                \
                & SoftMask++ (ours) & 4M &\textbf{80.9}  &  \textbf{80.6} & \textbf{80.6} & \textbf{81.6}  & \textbf{75.0} & \textbf{75.1} \\
                \bottomrule
            \end{tabular}
            }
	\end{center}
	\vspace{-4mm}
	\label{table:downstream}
\end{table}

\subsection{Results on Image-Text Retrieval}
\label{sub:itr}

Table~\ref{table:fine_tune_itr} presents the overall results of the proposed algorithm and other baselines on Flickr30K and MS-COCO for the fine-tuning setting, where our approach outperforms the baseline methods in most cases, especially for R@1.
For completeness, we provide the results from ALIGN \cite{jia2021scaling}, which uses 1.8B image-text pairs (1.2B unique images) for pretraining.
The results show that our model successfully transfer knowledge for the image-text retrieval task.

Table~\ref{table:zero_shot} illustrates the overall results of the proposed algorithm and other baselines in a zero-shot setting on the Flickr30K and MS-COCO datasets.
Our approach outperforms other methods in most experimental settings and also surpasses ALIGN~\cite{jia2021scaling}, which employs about 300$\times$ times more pretraining images.
The results illustrate that the proposed framework learns better generalizable representations than the compared methods.

\subsection{Results on Other Downstream Tasks}
\label{sub:downstream}
Table~\ref{table:downstream} summarizes the results of the proposed algorithm on the VE, NLVR, and VQA tasks. 
Among the methods without pretrained object detectors, our algorithm achieves state-of-the-art performance for all evaluation metrics.
Since the evaluated tasks employ different types of objective functions for training, the results show the generalizability of our multi-modal representation learning technique.
Another important feature of these tasks is that they require fine-grained reasoning about images and texts. 
The results demonstrate that the representations learned by the proposed soft masking approach and the other regularization techniques are effective for fine-grained recognition.
Note that VinVL~\cite{zhang2021vinvl} exploits about 6M images, which is larger than ours, to train the object detector before multi-modal representation learning.

\begin{table}[t]
	\centering
	\caption{Ablation study results of the image-text retrieval task on the MS-COCO dataset. 
	The baseline algorithm without all the three components, corresponding to the first row of this table, is the same as ALBEF~\cite{li2021align}.
	The bold-faced numbers indicate the best performance in each column.
	}
	\vspace{-2mm}
	\setlength\tabcolsep{8pt}
	\scalebox{0.85}{
		\begin{tabular}{ccccc}
			\toprule
			 \multicolumn{1}{c}{SoftMask} & \multicolumn{1}{c}{Focal ITC} & \multicolumn{1}{c}{MMDA}  & \multicolumn{1}{c}{TR@1}  & \multicolumn{1}{c}{IR@1}  \\
			\midrule
			&&  & 73.10  & 56.80  \\ \midrule
			$\surd$ &  & & 75.74  & 58.24   \\
			&  $\surd$& & 75.66 & 58.85   \\
			& &  $\surd$ & 75.26  & 58.85  \\  \midrule
			$\surd$  & $\surd$  && 76.06   &59.54   \\
			$\surd$ & & $\surd$  & 75.98  & 59.44  \\
			& $\surd$  & $\surd$  & 76.16   & 59.62   \\ \midrule
			$\surd$ &$\surd$ & $\surd$ &\textbf{76.62}  & \textbf{60.15}    \\
			\bottomrule
		\end{tabular}
	}
	\label{table:abl}
	\vspace{2mm}
\end{table}

\begin{figure*}[t]
	\centering
	\begin{subfigure}[t]{0.49\textwidth}
		\includegraphics[width=\textwidth]{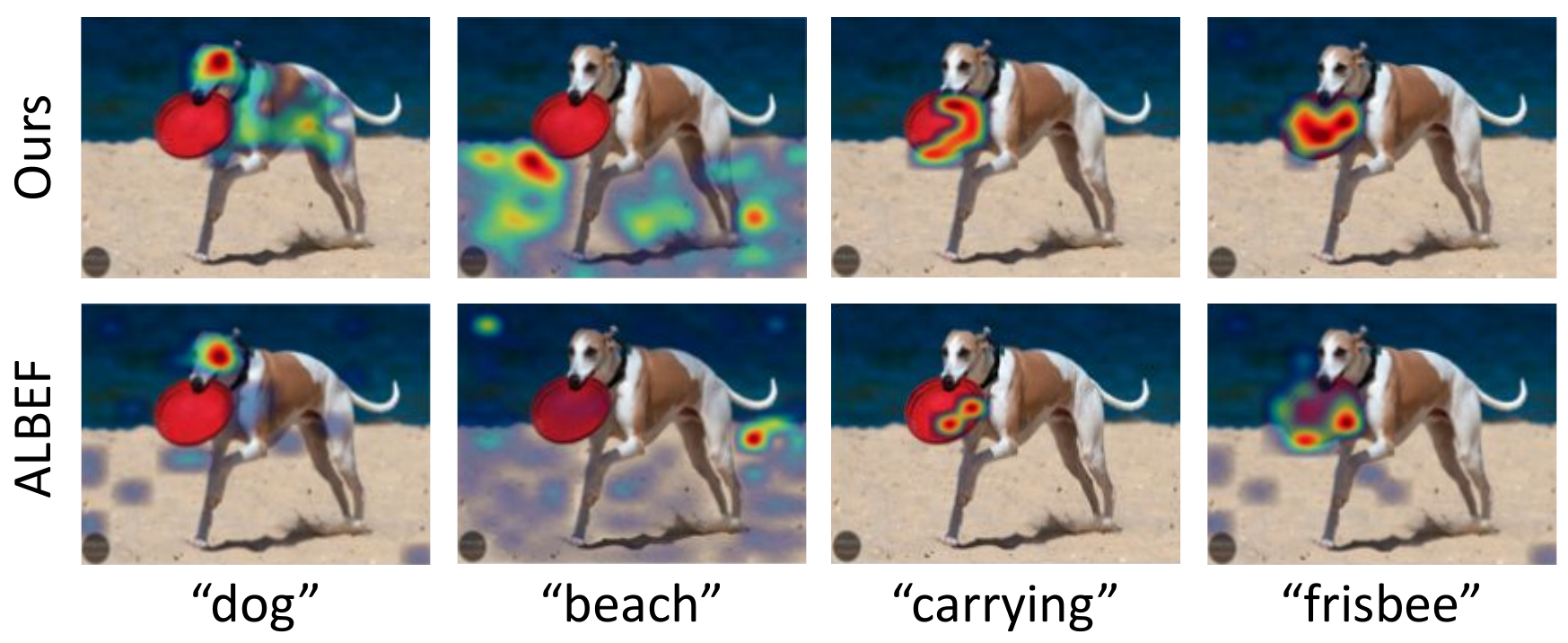} 
		\subcaption{A dog on a beach carrying a frisbee.}
	\end{subfigure}
	\begin{subfigure}[t]{0.49\textwidth}
		\includegraphics[width=\textwidth]{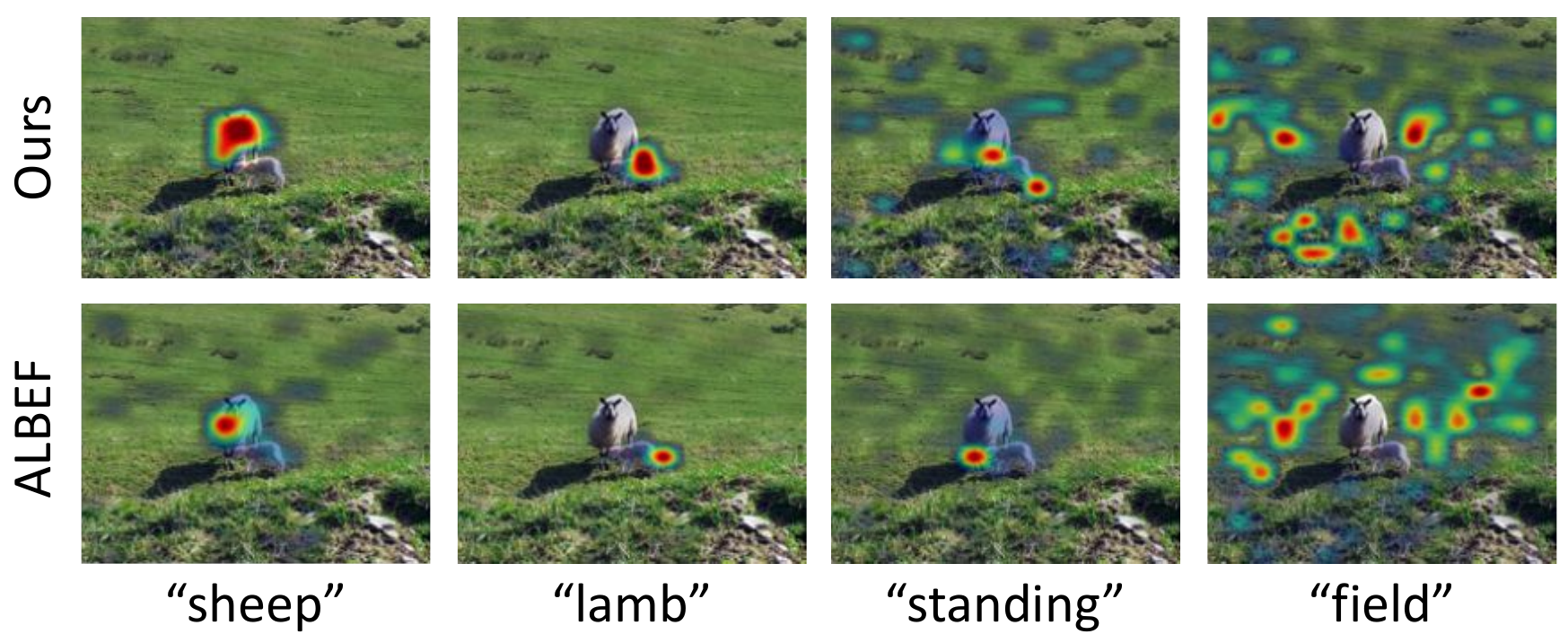}
		\subcaption{A sheep and lamb standing in the field.}
	\end{subfigure} 
	\vspace{-2mm}
	\caption{Word-conditional Grad-CAM visualization of our method and ALBEF~\cite{li2021align} after pretraining.	
	} 
	\vspace{-2mm}
\label{fig:qual}
\end{figure*}

\subsection{Ablation Study}
\label{sub:ablation}
We perform several ablation studies on the image-text retrieval task with the MS-COCO dataset and report R@1 performance to analyze the effectiveness of each component of our approach.
From the results in Table~\ref{table:abl}, we have the following observations.
First, our soft masking strategy enhances the overall transferability, by generating diverse visual features as we hypothesized.
Second, the proposed focal ITC loss boosts performance by large margins. 
These observations support our claim that concentrating on the hard examples and focusing less on easy ones alleviate overfitting while handling the class imbalance issue effectively.
Last, multi-modal data augmentation (MMDA) delivers strong regularization effect, indicating that adopting augmented examples with diverse aspects for vision-language pretraining has a positive effect.

\vspace{-3mm}
\paragraph{Soft vs. Random Masking}
Table~\ref{table:softrand} illustrates the comparison between our text-driven soft masking (SoftMask) and the random hard masking (RandMask) strategy.
For RandMask, we randomly remove regions in an image embedding with pre-defined ratios of 0.3 and 0.5.
SoftMask outperforms the hard RandMask, which is mainly because soft masking manages to populate diverse and, more importantly, semantically meaningful visual embeddings with varying rates given by Grad-CAM.

\begin{table}[t]
	\centering
	\caption{Image-text retrieval results on MS-COCO by varying masking strategy on visual features. 
	The bold-faced numbers indicate the best performance.
	}
	\setlength\tabcolsep{12pt}
	\scalebox{0.85}{
		\begin{tabular}{lcc}
			\toprule
			 Method  & \multicolumn{1}{c}{TR@1}  & \multicolumn{1}{c}{IR@1} \\
			\midrule
			RandMask ($p=0.3$) &75.82  &  59.52  \\
			RandMask ($p=0.5$) &76.22  &  59.74  \\ \midrule 
			SoftMask (Ours) &\textbf{76.62} &  \textbf{60.15}   \\
			\bottomrule
		\end{tabular}
	}
	\vspace{-2mm}
	\label{table:softrand}
\end{table}

\begin{table}[t]
	\centering
	\caption{
	Performance comparisons in the Image-text retrieval task on the MS-COCO dataset by applying SoftMask to the ITM and MLM pretraining tasks. Our default setting adopts SoftMask only for ITM.
	The bold-faced numbers indicate the best performance.
	}
	\setlength\tabcolsep{12pt}
	\scalebox{0.85}{
		\begin{tabular}{cccc}
			\toprule
			 $\mathcal{L}_\text{ITM}^{*}$ & $\mathcal{L}_\text{MLM}^{*}$ & \multicolumn{1}{c}{TR@1}  & \multicolumn{1}{c}{IR@1} \\
			\midrule
			&  & 76.16 & 59.62 \\
			$\surd$ & & \textbf{76.62} & \textbf{60.15} \\
			$\surd$ & $\surd$ & 76.21 & 59.84 \\
			\bottomrule
		\end{tabular}
	}
	\vspace{-2mm}
	\label{table:softtasks}
\end{table}

\vspace{-3mm}
\paragraph{SoftMask for MLM}
We conduct additional experiments to investigate whether our soft-masked visual features are also beneficial for the MLM task.
Table~\ref{table:softtasks} presents the results from the MLM loss with the soft-masked features, where we observe that SoftMask is not particularly effective for MLM and fails to improve performance  in downstream tasks according to our experiments.
This is partly because the reconstruction task is not well addressed when the signals from both modalities are imperfect. 
Meanwhile, ITM enjoys the regularization effect since the task depends more on global information in both modalities.

\subsection{Computational Complexity}
\label{sub:cost}
Two minor modifications of our backbone network, ALBEF~\cite{li2021align} affect computational cost.
First, we unify the forward path in the ITM and MLM tasks by using the same masked text input and reduce computation even with performance improvements.
Second, our soft masking strategy incurs an additional computation for computing Grad-CAM and feedforwarding soft-masked features through the fusion encoder.
Table~\ref{tab:comp_cost} presents the computational analysis compared to the ALBEF~\cite{li2021align} baseline.
On a single NVIDIA Quadro RTX GPU, ALBEF requires 1.94 seconds per iteration while our algorithm requires 1.98s with batch size 128.

\subsection{Qualitative Results}
\label{sub:qual}
Figure~\ref{fig:qual} illustrates qualitative comparisons between our algorithm and ALBEF~\cite{li2021align}, where we provide word-conditional Grad-CAM visualizations of two image-caption pairs from the MS-COCO dataset.
The followings are the lessons from the observation of the qualitative results.
First, our model provides more accurate and comprehensive coverage of the objects, which are relevant to the corresponding caption.
For instance, ALBEF only highlights the head of the dog, which is the most discriminative part, while our model is also activated on its body.
Second, our model obtains more accurate activations for action attributes.
For the action ``standing", ALBEF only highlights the feet of the sheep while ours recognizes the feet of both animals.
Last, our model may be biased toward certain objects or background as, for example, Grad-CAM of ``standing" fires on fields, if the most salient parts are heavily masked.

\begin{table}[!t]
\centering
\caption{Computation cost per a single pretraining iteration compared to ALBEF~\cite{li2021align}.
	We measure computational cost on a single NVIDIA Quadro RTX GPU with 128 batch size per GPU.}
	\setlength\tabcolsep{6pt}
	\scalebox{0.85}{
		\hspace{-0.2cm}
			\begin{tabular}{lcc}
			\toprule
			          Methods                                                   &  \multicolumn{1}{c}{Time (sec/it)}    & \multicolumn{1}{c}{GPU Mem. (GB/GPU)}  \\ \midrule
				  ALBEF~\cite{li2021align}                                       & 1.94  &   41.5     \\   
				  Ours w/o SoftMask & 1.91 & 38.7 \\  
				  SoftMask++ (ours)       & 1.98  &    41.4      \\ 
			\bottomrule
			\end{tabular} 
		}
	\vspace{-2mm}
	\label{tab:comp_cost}
\end{table}

\section{Conclusion}\label{sec:conclusion}	
In this paper, we presented a novel visual-linguistic representation learning framework based on the explainable soft feature masking strategy and the regularizations via diversifications.
We proposed a novel language-driven soft feature masking strategy to populate visual features with various aspects, where the most discriminative parts of an image are softly masked based on the stochastically estimated contribution of each local region to the matching score.
We also introduced a focal loss for the image-text contrastive learning objective, which addresses the inherent limitations of the overfitting and bias issues.
Last, we manipulate multi-modal data augmentation strategies for constructing more diversified samples by applying text maskings and rendering distortions on images.
Our algorithm achieves outstanding performance compared to existing vision-language pretraining methods on various vision-language downstream tasks.

\vspace{-3mm}
\paragraph{Acknowledgments}
This work was partly supported by Samsung SDS and the IITP grants [No.2022-0-00959, (Part 2) Few-Shot Learning of Causal Inference in Vision and Language for Decision Making, No.2021-0-01343, Artificial Intelligence Graduate School Program (Seoul National University), No.2021-0-02068, Artificial Intelligence Innovation Hub] funded by the Korean government (MSIT).

\clearpage
\appendix

\section*{Appendix}

\section{Datasets for Pretraining}
\label{sec:detail_pretrain}
We use the large-scale image-text corpora, which include MS-COCO~\cite{lin2014microsoft}, Visual Genome~\cite{krishna2017visual}, SBU Captions~\cite{ordonez2011im2text}, and Conceptual Captions~\cite{sharma2018conceptual} for pretraining for fair comparisons with the previous works~\cite{li2021align,yang2022vision,duan2022multi,chen2020uniter}.
Table~\ref{table:corpus_supp} presents the statistics of the pretraining datasets such as the number of images and captions.
Note that the size of the CC3M dataset is slightly different from other works~\cite{li2021align,yang2022vision} since CC3M  is webly-crowded and some of the download links are expired.
As a result, the pretraining corpora consist of 4M images and 5.1M image-text pairs.

\begin{table}[h!]
	\centering
	\caption{Statistics of pretraining datasets
	}
	\scalebox{0.9}{
		\begin{tabular}{lcccc}
			\toprule
			                    &MS-COCO  & VG      & SBU     & CC3M   \\ \midrule 
			\# Images    &113K            & 100K  & 858K    &  2.89M  \\ 
			\# Captions  &567K           & 769K   & 858K    & 2.89M   \\  
			\bottomrule
		\end{tabular}
	}
	\label{table:corpus_supp}
\end{table}

\begin{figure*}[t!]
	\centering
	\begin{subfigure}[t]{0.49\textwidth}
		\includegraphics[width=\textwidth]{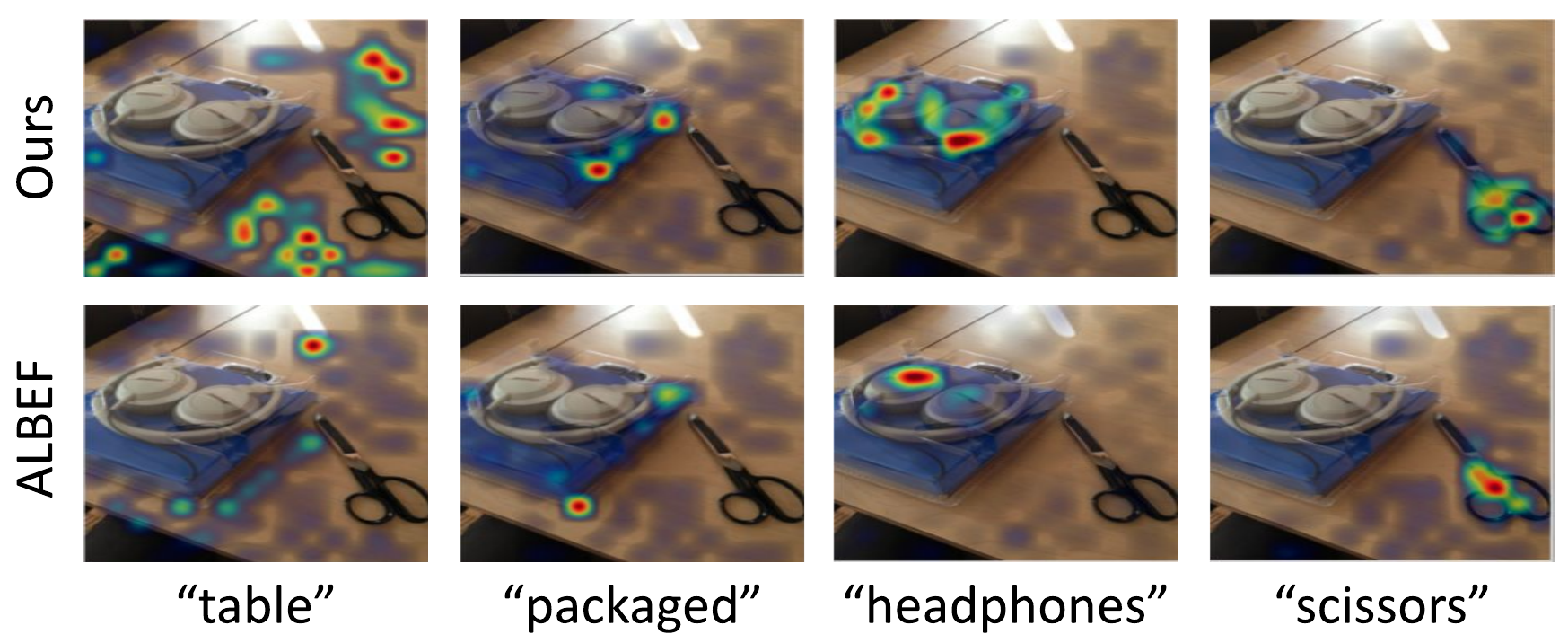}
		\subcaption{On the table is packaged headphones and scissors.}
	\end{subfigure}
	\begin{subfigure}[t]{0.49\textwidth}
		\includegraphics[width=\textwidth]{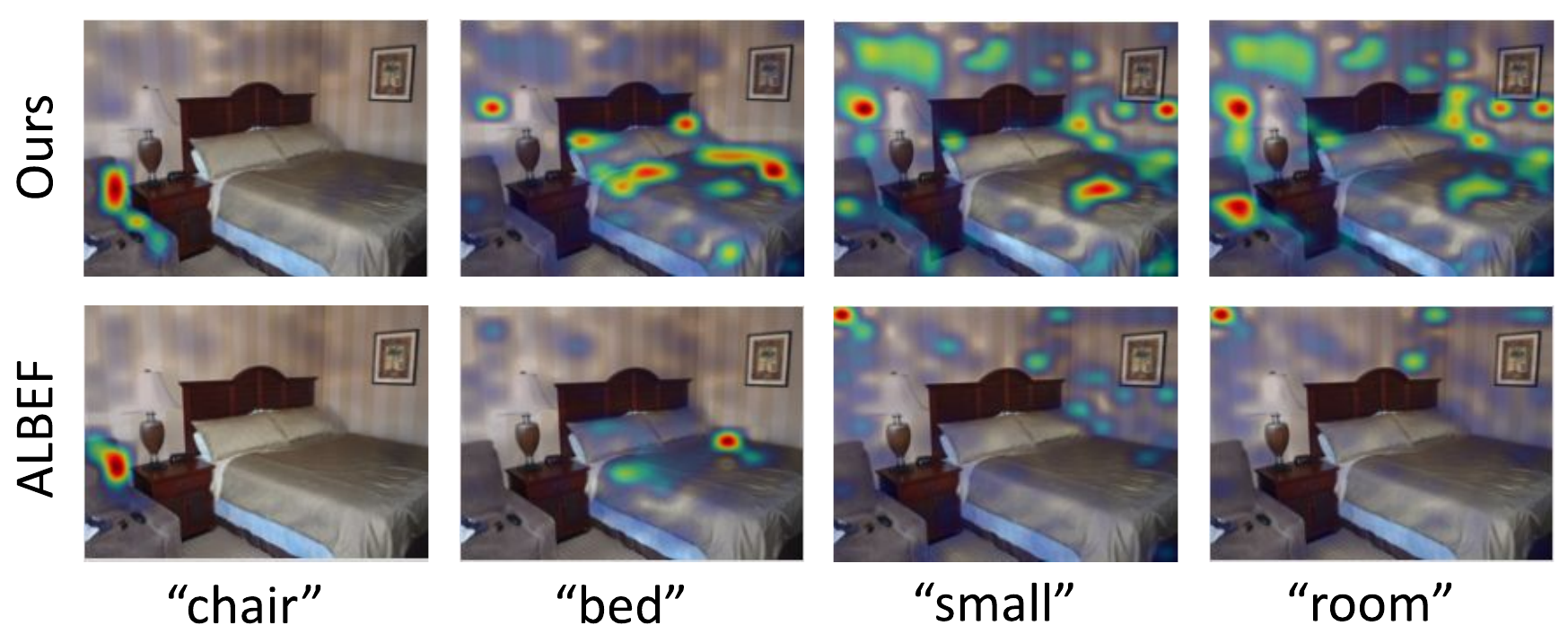} 
		\subcaption{A chair and a bed in a small room.}
	\end{subfigure} 
	\caption{Additional word-conditional Grad-CAM visualization of our method and ALBEF~\cite{li2021align} after pre-training.	
	} 
	\vspace{-0.3cm}
\label{fig:add_qual_supp}
\end{figure*}

\section{Implementation Details}
\label{sec:impl}
We employ ViT-B/16~\cite{dosovitskiy2021image} with 12 layers as our vision encoder and initialized it using the weights pretrained on ImageNet-1K~\cite{touvron2021training}. 
The text encoder and multi-modal encoder are initialized by the first 6 layers and the last 6 layers of BERT$_\text{base}$~\cite{devlin2019bert}, respectively.
We optimize our model using batch size of 1024 on 8 NVIDIA RTX A6000 GPUs for 30 epochs.
We adopt AdamW optimizer with weight decay 0.02.
The learning rate is initialized to 0.00002 and is warmed up to $2 \times 10^{-4}$ for 1,000 iterations.
After warm up, the learning rate is decayed to $2 \times 10^{-5}$ following the cosine scheduling.

In order to generate soft masks, we compensate the total weights of masked regions to meet one-half of the total number of patches.
By adopting this method, we can guarantee that the embedding of the image after being softly masked will not diverge too significantly from the original embedding, nor will it be excessively similar to it.

\section{Details on Downstream Tasks}
\label{sec:detail_down}

Below we provide implementation details of the vision-language downstream tasks to evaluate our pretraining approach, including Image-Text Retrieval (ITR), Visual Entailment (VE), Visual Question Answering (VQA), and Natural Language Visual Reasoning (NLVR). 
For all downstream tasks, we use  AdamW optimizer, RandAugment, cosine learning rate scheduling, and a weight decay, whose hyperparameters are the same as the pretraining procedure.

\paragraph{Image-Text Retrieval (ITR)}
For the ITR task, we conduct our experiments on the Karpathy split~\cite{karpathy2015deep} of the Flickr30K~\cite{plummer2015flickr30k} and MS-COCO~\cite{lin2014microsoft} datasets, which consist of 29k/1k/1k and 113k/5k/5k images for train/validation/test, respectively.
We fine-tune the model using the ITM and ITC losses, using a batch size of 256, with a learning rate of 0.00001, for 10 epoch for Flickr30K and 5 epochs for MS-COCO.
For inference, we first obtain top-$k$ candidates based on similarity scores from the uni-modal encoders, as shown in (\textcolor{red}{1}) of the main paper, and then compute their ITM scores given by (\textcolor{red}{4}) of the main paper, to rank the candidates, where $k$ is set to 128 for Flickr30K and 256 for MS-COCO.

\paragraph{Visual Entailment (VE)}
We use the SNLI-VE~\cite{xie2019visual} dataset, which consists of 30k/1k/1k images for train/validation/test, respectively.
The dataset is built upon the Flickr30K and Stanford Natural Language Inference (SNLI)~\cite{bowman2015large} datasets.
We optimize the pretrained model for 5 epochs using a batch size of 256 with a learning rate of 0.00002.

\paragraph{Natural Language Visual Reasoning (NLVR)}
We evaluate our model on the NLVR$^2$~\cite{suhr2019corpus} dataset, which consists of 86k/7k/7k examples for train/dev/test, respectively.
Since the task requires a pair of images as input, we modify our model by duplicating the transformer block, as mentioned in Section \textcolor{red}{5.2}.
We first train the pretrained model using a 4M pretraining corpus for one more epoch to adjust the modified model, using a batch size of 256 with a learning rate of 0.00002.
We fine-tuned the adjusted model for 10 epochs, using a batch size of 128 with a learning rate of 0.00002.

\paragraph{Visual Question Answering (VQA)}
We conduct experiments on VQA2.0 dataset~\cite{goyal2017making}, which is based on the images collected from the MS-COCO dataset.
The dataset consists of 83k/41k/81k images for train/validation/test, respectively.
Following the previous works~\cite{li2021align,yang2022vision}, we utilize both the training and validation sets for training, and also include additional question-answer pairs from the Visual Genome dataset.
Since each question in the VQA2.0 dataset is associated with 10 answers, we also weigh the loss for each answer based on its frequency among all answers, following~\cite{li2021align}.
We fine-tune the pretrained model for 8 epochs, using a batch size of 256, with a learning rate of 0.00002.

\section{Design Choice for SoftMask}
\label{sec:detail_down}
We conducted an additional experiment to compare Grad-CAM and normalized cross-attention maps to generate soft masks.
Table~\ref{table:ca} presents that utilizing Grad-CAM outperforms using normalized cross-attention maps.
It shows that Grad-CAM provides more suitable guidance than cross-attention maps to generate the soft mask.

\begin{table}[t!]
	\centering
	\caption{Image-text retrieval results on MS-COCO with Grad-CAM (ours) and normalized cross-attention map. 
	The bold-faced numbers indicate the best performance.
	}
	\scalebox{0.9}{
		\begin{tabular}{l|cc}
			\toprule
			 Method  & \multicolumn{1}{c}{TR@1}  & \multicolumn{1}{c}{IR@1} \\
			\hline\hline
			Cross-Attention &75.68  &  59.48  \\
			Grad-CAM (Ours) &\textbf{76.62} &  \textbf{60.15}   \\
			\bottomrule
		\end{tabular}
	}
	\label{table:ca}
\end{table}

\section{Additional Qualitative Results}
\label{sec:add_qual}
In Figure~\ref{fig:add_qual_supp}, we show more visualizations of word-conditional Grad-CAM of our method and ALBEF~\cite{li2021align} after pre-training.
In general, our model provides Grad-CAM with more accurate and diverse attributes of the concepts than ALBEF, which is also presented in Figure 3 in the main paper. 
However, our model may learn bias towards the object and scene if the most discriminative part is masked with high weights, as we discussed in Section 5.7.
For instance, in Figure~\ref{fig:add_qual_supp} (b) our model fires on the wall for ``bed'', since bed usually comes with wall.

{\small
\bibliographystyle{ieee_fullname}
\bibliography{egbib}
}

\end{document}